\documentclass[letterpaper, 10 pt,conference]{IEEEtran}
\IEEEoverridecommandlockouts
\usepackage{cite}
\usepackage{amsmath,amssymb,amsfonts}
\usepackage{graphicx}
\usepackage{textcomp}
\usepackage{xcolor}
\def\BibTeX{{\rm B\kern-.05em{\sc i\kern-.025em b}\kern-.08em
    T\kern-.1667em\lower.7ex\hbox{E}\kern-.125emX}}
\usepackage{multirow}
\usepackage{subfigure}

\usepackage{amsmath}
\usepackage{graphicx}
\usepackage{algorithm}
\usepackage[noend]{algpseudocode}
\usepackage{color}
\usepackage{tabularray}
\usepackage{array}    
\usepackage{soul}
\usepackage{booktabs}

\begin{document}

\title{\LARGE \bf
\vspace{0.55cm}
T2S: Tokenized Skill Scaling for Lifelong Imitation Learning
}


\author{Hongquan Zhang$^{1}$$^{2}$, Jingyu Gong$^{1}$, Zhizhong Zhang$^{1}$, Xin Tan$^{1}$, Yanyun Qu$^{3}$, Yuan Xie$^{*}$$^{1}$$^{2}$
\thanks{
\newline $^{*}$Corresponding author
\newline $^{1}$School of Computer Science and Technology, East China Normal University, Shanghai 200062, China. (e-mails: 52285901010@stu.ecnu.edu.cn)
\newline $^{2}$Shanghai Innovation Institute, Shanghai 200062, China.
\newline $^{3}$School of Informatics, Xiamen University, Xiamen, Fujian 361005, China.
}
}

\maketitle

\begin{abstract}
The main challenge in lifelong imitation learning lies in the balance between mitigating catastrophic forgetting of previous skills while maintaining sufficient capacity for acquiring new ones. However, current approaches typically address these aspects in isolation, overlooking their internal correlation in lifelong skill acquisition. We address this limitation with a unified framework named Tokenized Skill Scaling (T2S).  Specifically, by tokenizing the model parameters, the linear parameter mapping of the traditional transformer is transformed into cross-attention between input and learnable tokens, thereby enhancing model scalability through the easy extension of new tokens. Additionally, we introduce language-guided skill scaling to transfer knowledge across tasks efficiently and avoid linearly growing parameters. Extensive experiments across diverse tasks demonstrate that T2S: 1) effectively prevents catastrophic forgetting (achieving an average NBT of \(1.0\%\) across the three LIBERO task suites), 2) excels in new skill scaling with minimal increases in trainable parameters (needing only \(8.0\%\) trainable tokens in an average of lifelong tasks), and 3) enables efficient knowledge transfer between tasks (achieving an average FWT of \(77.7\%\) across the three LIBERO task suites), offering a promising solution for lifelong imitation learning.
\end{abstract}


\section{Introduction}
\label{section_introduction}
Imitation learning enables robots to acquire complex task skills by observing and mimicking expert demonstrations, facilitating the deployment of robust strategies across various sequential decision-making tasks \cite{stepputtis2020language,xie2024decomposing,schaal1996learning,osa2018algorithmic,gavenski2024imitation}. This approach is particularly well-suited for scenarios where designing reward functions is challenging \cite{christiano2017deep,lee2019composing,leike2018scalable}, as well as for complex tasks where trial-and-error learning is too costly or poses safety risks \cite{gu2024review,garcia2015comprehensive}.

\begin{figure}[t!]
    \centering
    \includegraphics[width=0.9\linewidth]{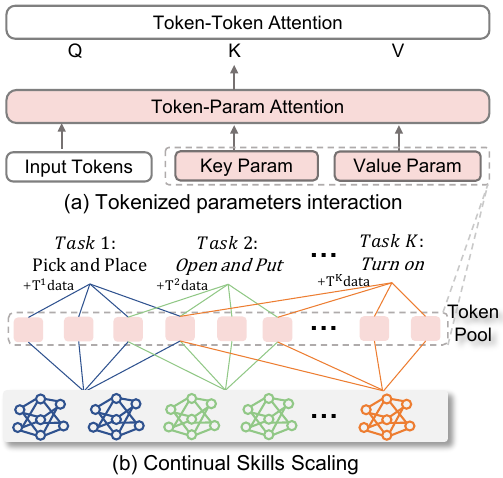}
    \caption{(a): We convert the traditional linear mapping of parameters into tokenized interactions to improve the scalability of the model through token extensions. (b): Through token sharing between tasks, we achieve efficient utilization of parameters while skills are continuously scaling.}
    \label{fig1}
\end{figure}

However, most existing research lacks the ability to support incremental skill acquisition and cannot facilitate lifelong skill adaptation or the open-ended expansion of skill scaling \cite{liang2024never}.
This drawback conflicts with the dynamic and consequent demands of real-world task flows, \textit{i.e.},  when robots need to expand their skills incrementally, traditional frameworks often face the dual challenges of catastrophic forgetting and skill capacity saturation \cite{lesort2020continual}. The former arises from the parameter-sharing mechanism in neural networks, where the acquisition of new skills inevitably interfere with learned knowledge, akin to the memory conflict phenomenon in connectionist networks as described by McCloskey et al. \cite{mccloskey1989catastrophic} in cognitive science. 
The latter is due to the loss of plasticity in learning from extended training on new data in a single neural network \cite{dohare2024loss}.
Consequently, constructing lifelong imitation learning frameworks capable of simultaneously maintaining knowledge retention and architectural scalability has become a central challenge for promoting robots' adaptation to open environments.

LOTUS \cite{wan2024lotus} employs hierarchical skill learning to acquire new skills continuously and integrates them into a dynamic skills library, effectively improve scalability but still struggle with catastrophic forgetting. M2Distill \cite{roy2024m2distill} addresses catastrophic forgetting by transferring multi-modal knowledge from the teacher model to the student model using knowledge distillation techniques.
However, this approach lacks scalability and suffers from reduced plasticity in lifelong learning, thereby limiting the model’s ability to acquire new skills over time.
Despite recent advancements in lifelong imitation learning, existing methods predominantly focus on either skill capacity saturation or catastrophic forgetting in isolation, overlooking the internal correlation of these challenges in lifelong skill acquisition.

To address the limitations of current methods, we propose a \textbf{T}okenized \textbf{S}kill \textbf{S}caling (T2S) approach. 
Specifically, we transform the traditional linear parameter mapping in the Transformer architecture with a cross-attention mechanism based on token-parameter interactions, as illustrated in Fig. \ref{fig1}(a). This full parameter tokenization enables the acquisition of new skills by simply extending additional tokens, thereby addressing the loss of model plasticity commonly encountered in lifelong learning. However, naïvely expanding new tokens for each task results in a linear increase in model parameters, leading to significant storage overhead. To mitigate this, we introduce a token pool, from which relevant tokens are selected based on task descriptions, as depicted in Fig. \ref{fig1}(b). Tokens shared across tasks can be seen as tokenized atomic skills. When new tasks are introduced, task descriptions retrieve a subset of atomic skills from the token pool and combine them with newly added tokens to acquire task-specific skills. With the progressive acquisition of skills, the number of atomic skills in the token pool increases, thereby providing more shared knowledge for learning subsequent skills. This mechanism promotes inter-task skill sharing, allowing for the efficient acquisition of new skills with only a modest increase in tokens, while effectively controlling the growth of model parameters.
In summary, our main contributions are as follows:
\begin{itemize}
    \item We introduce a scalable lifelong imitation learning framework that enhances model scalability by tokenizing all learnable parameters.
    \item We propose a language-guided token activation and selection strategy that effectively transfers shared knowledge across tasks while mitigating the linear growth of model parameters.
    \item Our proposed method achieves state-of-the-art performance in mitigating catastrophic forgetting.
\end{itemize}


\section{Related Work}
\label{section_related_work}
\subsection{Lifelong Imitation Learning}
Lifelong learning has made significant progress in various fields, including computer vision \cite{zhang2024learning}, natural language processing \cite{huang2023knowledge}, multi-modal large language models \cite{zheng2023preventing}, and robotics \cite{kim2024online,meng2025preserving}. This work focuses on lifelong imitation learning, which enables robots to continuously acquire new skills through human demonstrations without forgetting previously learned skills. In recent years, numerous researchers have investigated this field. ER \cite{chaudhry2019tiny} mitigates catastrophic forgetting by retaining a portion of the demonstration trajectory for each task. EWC\cite{kirkpatrick2017overcoming} prevents forgetting by calculating the importance of model parameters and constraining updates to parameters critical to previous tasks while learning new ones. BUDS \cite{zhu2022bottom} employs a bottom-up approach to autonomously identify and organize skills from unsegmented, long-duration demonstration data, allowing robots to effectively manage complex and prolonged manipulation tasks. LOTUS \cite{wan2024lotus} uses open-vocabulary visual models for skill discovery and meta-controllers for skill integration, enabling robots to learn and adapt to new tasks continuously. M2Distill \cite{roy2024m2distill} addresses catastrophic forgetting by utilizing multimodal knowledge distillation to maintain consistency in the potential space across visual, verbal, and motor distributions during skill learning. In contrast, T2S enhances model scalability by tokenizing parameters and employing task descriptions to guide inter-task token share, allowing robots to learn new skills throughout their lifetime without being limited by linearly growing parameters and catastrophic forgetting.


\subsection{Parameter Scaling of Transformer}\label{sec2-pattention}
In recent years, with the rapid advancement of large language models such as GPT \cite{brown2020language}, DeepSeek \cite{bi2024deepseek}, and LLaMA \cite{touvron2023llama}, researchers have increasingly focused on efficiently scaling models from small to large sizes, driven by the substantial overhead of intensive training.
Gong et al. \cite{gong2019efficient} propose a stacking algorithm that transfers knowledge from a shallow model to a deeper one by progressively applying stacking to accelerate BERT \cite{devlin2019bert} training. LiGO \cite{wang2023learning} tackles the challenge of efficiently scaling pre-trained transformers by learning parameter growth through a factorized approach, decomposing the linear transformation into Kronecker-factorized width- and depth-growth operators. bert2BERT \cite{chen2022bert2bert} facilitates knowledge transfer from small models via function-preserving parameter initialization and accelerates training using a two-stage strategy. Tokenformer \cite{wang2025tokenformer} enhances architectural flexibility by leveraging the attention mechanism not only for computations among input tokens but also for interactions between tokens and model parameters. In this work, inspired by Tokenformer's high extensibility, we adopt its parameter tokenization approach as the foundation for lifelong imitation learning.

\section{Method}
\label{section_methodology}
Our method builds a behavior cloning network based on the token-Parameter attention (Pattention) layer, augmented by a tailored training and deployment strategy for lifelong learning in robotics. In subsequent sections, we begin with a problem formulation for lifelong imitation learning. We then present the core architecture of the Pattention-based behavior cloning model, emphasizing its scalability and flexibility. Finally, we introduce Language-Guided Skill Scaling—a novel token-sharing and expansion mechanism developed to fully harness the model’s potential in lifelong imitation learning.  
\begin{figure*}
    \centering
    \includegraphics[width=0.9\linewidth]{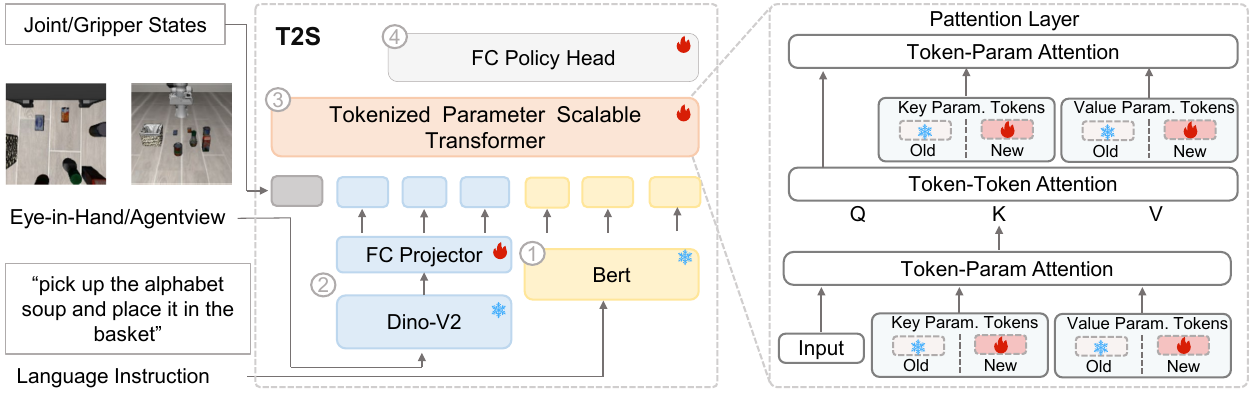}
    \caption{An overview of T2S: On the left is the full pipeline of our model architecture, and on the right is the Pattention layer in the Tokenized Parameter Scalable Transformer.}
    \label{motivation}
\end{figure*}
\subsection{Problem Formulation}
Lifelong Robot Learning constitutes a fundamental aspect of embodied intelligence, distinguished by its capability to incrementally acquire and refine skills through continuous engagement with a sequence of learning tasks \( \{T^1, \dots, T^K\} \).
This foundational robot learning paradigm can be formulated using a finite-horizon Markov Decision Process (MDP), denoted as \( M = (S, A, \mathcal{T}, H, \mu_0, R) \). Here, \( S \) represents the state space, \( A \) denotes the action space, \( \mathcal{T}: S \times A \to S \) specifies the transition function, \( H \) is the length of each task episode, \( \mu_0 \) characterizes the initial state distribution, and \( R: S \times A \to \mathbb{R} \) is the reward function. Due to the sparsity of \(R\) commonly encountered in robotic learning scenarios, a binary goal predicate \( g: S \to \{0,1\} \) is employed as a replacement to explicitly indicate the achievement of task objectives. Within the lifelong learning context, the robot is tasked with developing a unified policy \( \pi \) capable of sequentially adapting to the specific demands of each task \( T^k \), which is uniquely characterized by its initial state distribution \( \mu_0^k \) and a task-specific goal predicate \( g^k \). In this paradigm, we assume \( S \), \( A \), \( \mathcal{T}\), and \( H \) are consistent across all tasks. After sequentially experiencing tasks up to task \( k \), the robot aims to optimize its policy to maximize the expected return, formulated as:
\begin{equation}\label{eq:lrl}
    \max_{\pi} J(\pi) = \frac{1}{K}\sum_{p=1}^{K}\mathbb{E}_{s_t^p,a_t^p \sim \pi(\cdot;T^p),\, \mu_0^p}\left[\sum_{t=1}^{H}g^p(s_t^p)\right].
\end{equation}

\textbf{Lifelong Imitation Learning.} Given the inherent challenges associated with sparse-reward reinforcement learning, we adopt a practical scenario in which a user provides a small set of demonstrations for each task within a sequence. Specifically, we consider a lifelong imitation learning scenario, in which each task \( T^k \) is associated with \( N \) expert demonstrations \( D_k = \{\tau_i^k\}_{i=1}^{N} \) and corresponding natural language task instruction \( l^k \). Each demonstration trajectory \( \tau_i^k \) consists of observation-action pairs, formally defined as \( \tau_i^k = \{(o_t, a_t)\}_{t=0}^{L^k} \), where \( L^k \leq H \). The observation \( o_t \) comprises the robot's sensory inputs, including perceptual observations and proprioceptive data from the robot’s joints and gripper.  
In practice, due to partial observability in MDPs \cite{hausknecht2015deep}, observations \( o_t \) alone do not fully satisfy the Markov property. Thus, consistent with prior work, we define the state \( s_t \) as the historical sequence of observations up to time \( t \), formally represented as: \(s_t \equiv o_{\leq t} \triangleq \{(o_t, a_t)\}_{t=0}^{L^k}.\)This formulation aligns with the lifelong imitation learning setting described by LIBERO \cite{liu2023libero}, whose objective remains consistent with the one introduced in Eq. \ref{eq:lrl}. However, during training, behavioral cloning \cite{bain1995framework} is employed as a surrogate objective, defined as follows:
\begin{equation}\label{eq:lbc}
    \min_{\pi} J(\pi) = \frac{1}{K}\sum_{k=1}^{K}\mathbb{E}_{(o_t,a_t)\sim D_k}\left[\sum_{t=1}^{L^k}-\log \pi(a_t|o_{\leq t}; T^k)\right].
\end{equation}

\subsection{Pattention-Based Behavior Cloning}\label{sec3-2}
The architecture of T2S consists of four main components, as illustrated in Fig. \ref{motivation}: 1) A frozen Large Language Model (LLM) that maps the task instruction into language tokens, 2) a frozen visual encoder that transforms visual inputs into a set of tokens and subsequently projects them into the same dimensional space as the language tokens, 3) a token-scalable transformer that continuously encodes these tokens into action tokens, and 4) a policy head that decodes the action tokens into robot-executable actions. During training, the model is trained end-to-end using a mean squared error objective on the demonstration data. We take a Tokenized Parameter Scalable Transformer (TPST) as the basic block and build the network in LIBERO manner \cite{liu2023libero}. This design provide a fundamental advantage for lifelong imitation learning.

\begin{figure*}[t]
    \centering
    \includegraphics[width=0.9\linewidth]{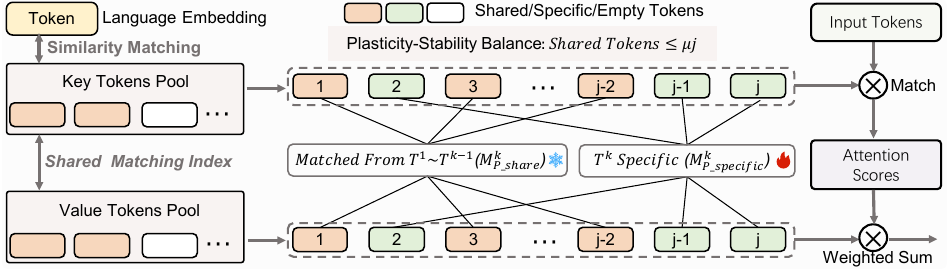}
    \caption{The visual explanation of Language-Guided Skill Scaling of task \(k (k > 1)\).}
    \label{pipeline}
\end{figure*}

In this work, we build the TPST upon the Tokenformer \cite{wang2025tokenformer}. Tokenformer operates through token-parameter attention (Pattention) layers, wherein a set of learnable tokens are treated as model parameters. These tokens interact with input tokens through cross-attention mechanisms, allowing the model to capture the relationships between the input and parameter tokens. For each TPST block, the input and output tokens are marked as \( \mathcal{I} \in \mathbb{R}^{T \times d_1} \) and \( \mathcal{O} \in \mathbb{R}^{T \times d_2} \) respectively. Here, \( T \) denotes the sequence length, and \( d_1 \) and \( d_2 \) are the input and output dimensions. Two sets of \( n \) learnable parameter tokens are introduced: \( K_P \in \mathbb{R}^{n \times d_1} \) for the keys and \( V_P \in \mathbb{R}^{n \times d_2} \) for the values. The output \( \mathcal{O} \) from the scaled dot-product attention mechanism is computed as follows:
\begin{equation}\label{pattention}
    \text{Pattention}(X, K_P, V_P) = \Theta \left( X \cdot K_P^{\top} \right) \cdot V_P,
\end{equation}
where \( \Theta \) denotes the softmax operation. We adopt Pattention in place of the traditional Multi-Head Attention (MHA) and Feed-Forward Networks (FFN) in the transformer due to the flexible design of the Pattention layer, which facilitates the faster incremental development of larger models while reusing parameters from smaller, pre-trained counterparts.
However, despite its advantages in incremental training, Pattention is insufficient for lifelong imitation learning, arising from its linear growing tokens and inability to 
maintain knowledge from previously acquired skills when scaling to new tasks,
kindly refer to the answer to question 3 (Sec. \ref{subsec:ablation}) for experimental proofs.
\subsection{Language-Guided Skill Scaling}
A straightforward solution to address the drawbacks of Pattention in lifelong imitation learning would be to learn independent sets of tokens for each task and identify them via task IDs. However, this approach leads to a linear increase in model parameters as the number of tasks grows, resulting in significant storage overhead. To overcome this limitation, we propose a language-guided parameter token activation and expansion approach(shown in Fig. \ref{pipeline}). Such a sharing mechanism enables continuous scaling of new skills with much less extra storage (validated in Sec.~\ref{subsec:ablation}). Specifically, within each Pattention layer, we construct token pools \( KP \in \mathbb{R}^{n \times d} \) and \( VP \in \mathbb{R}^{n \times d} \) separately for keys and values, where \( n \) denotes the number of tokens and \( d \) represents the embedding dimension. For each task, the amount of tokens is \(j = n/K\).
These token pools are managed by a unified global mask \( M_G \in \{0,1\}^{n} \), indicating whether each token is utilized. A language embedding \( e^k \) is further leveraged to identify and select the most relevant tokens as follows:
\begin{equation}\label{token-match}
   M_{P}^k = \text{Top-}\mathcal{K}(<e^k, KP>,j),
\end{equation}
where \( <,> \) denotes the cosine similarity and \( M_{P}^k \in \{0,1\}^n \) is the mask of selected tokens for task \(k\). Particularly, the \(M_{P}^{k}\) is shared between \(KP\) and \(VP\). For the first task, all the selected tokens are trained as task-specific tokens, while in the following lifelong tasks, the selected tokens consist of two components: 1) task-shared tokens \( M^k_{P\_share} = M_{P}^k \wedge M_G \), which are reused from previous tasks, and 2) task-specific tokens \(M^k_{P\_specific} = M_{P}^k \wedge \neg(M_{P}^k \wedge M_G) \), which are newly initialized. To ensure that the new policy contains a sufficient number of parameter tokens, we introduce a hyper-parameter \(\mu\) to regulate the number of shared tokens. Specifically, the number of shared tokens is constrained to be at most \(\lfloor \mu \times j \rfloor\), where \(\mu \in [0,1]\). Finally, the selected key and value tokens interact with the input as defined in Eq. \ref{pattention}.

The entire training and evaluation procedure is presented in Algorithm \ref{algorithm:LGPS}; we omit the backpropagation process for simplicity. During the training stage (lines 6–23), all selected tokens are trained for the first task since no prior reference knowledge is available. This step serves as the initialization for the entire lifelong learning process, providing prior knowledge for subsequent tasks. In later tasks, shared knowledge can be extracted from previously learned tokens. As lifelong learning progresses, the token pool gradually accumulates knowledge, enabling subsequent tasks to achieve their objectives by training only a small number of task-specific tokens. During the evaluation stage (lines 25–28), all previously learned tasks are evaluated, and the mask \(M_P^i\) is applied to select corresponding tokens from \(KP\) and \(KV\) for each task \(i\).

\begin{algorithm}[h]
    \caption{Overall training and evaluation procedure of Language-Guided Skill Scaling}
    \label{algorithm:LGPS}
    \begin{algorithmic}[1]
    
    \State {\textbf{Input:} Token Pools \(KP \) and \(VP\) , global mask \(M_G\), task mask list \(M^K_P\), hyper-parameters \(\mu\), the number of tasks \(K\)}, language embedding \(e\), input \( \mathcal{I} \),  demonstration \(D\)
    \State {\textbf{Output:} Interacted Tokens \( O\)}
    \State {Random initialize \(KP\) and \(VP\), initialize \(M_G\) and \(M^K_p\) to all \(0\)}
    \For {k=1 to K}
        \State{\textbf{Training}}
        \State{\(M_{P}^k \leftarrow e^k,KP\)} \Comment{Via Eq.\ref{token-match} } 
        \If{\(k \neq 1\)}
                \State{\( M^k_{P\_sh} \leftarrow M_{P}^k \wedge M_G \)}
                \Comment{Shared}
                \If{sum(\( M^k_{P\_sh}\)) \(>  \lfloor \mu \times j \rfloor\)}
                    \State{\( M^k_{P\_sh} \leftarrow \) Reduce tokens until to \( \lfloor \mu \times j \rfloor\)}
                \EndIf

                \State{\(M^k_{P} \leftarrow\) 
                \textit{update} \(M_{P}^k\)  \textit{via} \(M^k_{P\_sh} \)} 
                \State{\(M^k_{P\_sp} \leftarrow M_{P}^k \wedge \neg(M_{P}^k \wedge M_G) \)}  
                \Comment{Specific}
        \EndIf
        \State{$M_G \leftarrow M_G \mid M_{P}^k$, $M_P^K \leftarrow M_{P}^k$} 

        \For{ epoch=1 to epochs}
            \If{k=1} 
                 \State{\(K,V \leftarrow KP \times M_{P}^{k}, KV \times M_{P}^{k}\),}
                 \State{$ \textit{S} \leftarrow \Theta (\mathcal{I} \cdot K^\top)$, $O \leftarrow S\cdot V$} 
            \Else
                \State{\(K_{sh}, V_{sh} \leftarrow KP  \times M^k_{P\_sh}, KV \times M^k_{P\_sh}\)}
                
                \State{\(K_{sp}, V_{sp} \leftarrow KP \times M^k_{P\_sp}, KV \times M^k_{P\_sp}\)}
                
                \State{ \( \textit{S} \leftarrow \Theta (\mathcal{I} \cdot K_{sh}^\top \textit{(detached)} + \mathcal{I} \cdot K_{sp}^\top) \)  }  

                \State{ \( O \leftarrow \textit{S} \cdot V_{sh} \textit{(detached)} + \textit{S} \cdot  V_{sp} \)  } 
            \EndIf
        \EndFor
        \State{\textbf{Evaluation}}
        \For{i=1 to k} \Comment{Ealuation all tasks before task k}
            \State{\(M_P^i \leftarrow M_P^K,i\)}
            \State{\(K,V \leftarrow KP \times M_{P}^{i}, KV \times M_{P}^{i}\)}
            \State{\( O \leftarrow \mathcal{I},K,V\)} \Comment{Via Eq. \ref{pattention}}
        \EndFor
    \EndFor
    \end{algorithmic}
\end{algorithm}
\section{Experimental Evaluation}
\label{section_experimental_evaluation}
\begin{table*}[h] 
\caption{Performance of the proposed method compared with state-of-the-art,  corresponding results are borrowed from M2Distill and LIBERO. The reported values are averages from three seeds, including the mean and standard error. All metrics are measured based on success rates $(\%)$.}
\centering
\setlength{\tabcolsep}{2pt} 
\resizebox{0.95\textwidth}{!}{
\begin{tblr}{
  colsep=2pt,
  cells = {c},
  cell{1}{1} = {r=2}{},
  cell{1}{2} = {c=3}{},
  cell{1}{5} = {c=3}{},
  cell{1}{8} = {c=3}{},
  vline{5,6} = {1}{},
  vline{8,6} = {1}{},
  vline{5,8} = {2}{},
  hline{1,3,9,10} = {-}{},
  hline{2} = {2-10}{},
}
Methods     & LIBERO-OBJECT &               &               & LIBERO-GOAL   &               &               & LIBERO-SPATIAL &               &               \\
           & FWT$(\nearrow)$ & NBT$(\searrow)$ & AUC$(\nearrow)$ & FWT$(\nearrow)$ & NBT$(\searrow)$ & AUC$(\nearrow)$ & FWT$(\nearrow)$  & NBT$(\searrow)$ & AUC$(\nearrow)$ \\
SEQUENTIAL &62.0 ($\pm$ 1.0) &63.0 ($\pm$ 2.0) &30.0 ($\pm$ 1.0) &55.0 ($\pm$ 1.0) &70.0 ($\pm$ 1.0) &23.0 ($\pm$ 1.0) &72.0 ($\pm$ 1.0) &81.0 ($\pm$ 1.0) &20.0 ($\pm$ 1.0) \\

EWC \cite{kirkpatrick2017overcoming} &56.0 ($\pm$ 3.0) &69.0 ($\pm$ 2.0) &16.0 ($\pm$ 2.0) &32.0 ($\pm$ 2.0) &48.0 ($\pm$ 3.0) &6.0 ($\pm$ 1.0) &23.0 ($\pm$ 1.0) &33.0 ($\pm$ 1.0) &6.0 ($\pm$ 1.0) \\

ER \cite{chaudhry2019tiny} &56.0 ($\pm$ 1.0) &24.0 ($\pm$ 1.0) &49.0 ($\pm$ 1.0) &53.0 ($\pm$ 1.0) &36.0 ($\pm$ 1.0) &47.0 ($\pm$ 2.0) &65.0 ($\pm$ 3.0) &27.0 ($\pm$ 3.0) &56.0 ($\pm$ 1.0) \\


LOTUS \cite{wan2024lotus} &74.0 ($\pm$ 3.0) &11.0 ($\pm$ 1.0) &65.0 ($\pm$ 3.0) &61.0 ($\pm$ 1.0) &30.0 ($\pm$ 1.0) &56.0 ($\pm$ 1.0) &- &- &- \\

M2Distill \cite{roy2024m2distill} &\underline{75.0 }($\pm$ 3.0) &\underline{8.0} ($\pm$ 5.0) &\textbf{69.0 ($\pm$ 4.0)} &\underline{71.0} ($\pm$ 1.0) &20.0 ($\pm$ 3.0) &57.0 ($\pm$ 2.0) &\underline{74.0} ($\pm$ 1.0) &11.0 ($\pm$ 1.0) &61.0 ($\pm$ 2.0)  \\
T2S      &\textbf{75.0 ($\pm$ 2.0)} &\textbf{4.0 ($\pm$ 1.0)} &\underline {66.0} ($\pm$ 2.0)  &\textbf{84.0 ($\pm$ 1.0)} &\textbf{-1.0 ($\pm$ 2.0)} &\textbf{70.0 ($\pm$ 2.0)} &\textbf{74.0 ($\pm$ 1.0)} &\textbf{0.0 ($\pm$ 1.0)}  &\textbf{63.0 ($\pm$ 2.0) }             
\end{tblr}}
\label{experiment:main}
\end{table*}

\subsection{Experimental Setup}
\textbf{Simulation Experiments.} We conduct experiments in simulation scenarios using the task suites from the lifelong robot learning benchmark LIBERO \cite{liu2023libero}. Specifically, we select three suites, namely LIBERO-OBJECT (10 tasks), LIBERO-GOAL (10 tasks), and LIBERO-SPATIAL (10 tasks). These benchmarks evaluate the robot's ability to transfer knowledge between different objects (declarative knowledge) and various instructions (procedural knowledge), respectively. Additional implementation details are provided in the supplementary materials.


\textbf{Evaluation Metrics.} To systematically evaluate the effectiveness of lifelong imitation learning methods for robotic manipulation tasks, we employ three standardized evaluation metrics: Forward Transfer (FWT), Negative Backward Transfer (NBT), and Area Under the Success Rate Curve (AUC), in alignment with prior research \cite{wan2024lotus,roy2024m2distill}. These metrics are based on task success rates across training procedures, providing a more accurate and meaningful representation of actual manipulation performance. Specifically, FWT quantifies the policy's adaptability and its ability to generalize effectively to new tasks, higher FWT scores indicate improved learning efficiency and successful transfer of prior knowledge. In T2S, we evaluate it after one epoch adaptation. In contrast, NBT evaluates how well the policy maintains knowledge acquired in previous tasks when exposed to new tasks, where lower values signify better retention and less interference. Lastly, AUC aggregates task success rates over the entire sequence of encountered tasks, serving as a comprehensive indicator of lifelong learning performance. Higher AUC values represent sustained effectiveness and robust task success throughout the robot's operational lifetime.
Denote $r_{i,j}$ as the agent’s performance on task $j$ immediately after learning from the first $i$tasks. These metrics are defined as follows:
$FWT = \sum_{m \in [M]} \frac{r_{m,m}}{M},NBT = \sum_{m \in [M]} \frac{NBT_m}{M},$ and $AUC = \sum_{m \in [M]} \frac{AUC_m}{M},$
where $NBT_m = \frac{1}{M - m} \sum_{q = m+1}^{M} \left( r_{m,m} - r_{q,m} \right),AUC_m = \frac{1}{M - m + 1} 
\left( r_{m,m} + \sum_{q = m+1}^{M} r_{q,m} \right).$

\subsection{Compared Methods}
We compare our method against the following baselines: 
\begin{itemize}
    \item \textbf{SEQUENTIAL}, which naively fine-tunes new tasks sequentially using the ResNet-Transformer architecture from LIBERO.
    \item \textbf{EWC} \cite{kirkpatrick2017overcoming}, a regularization-based continual learning approach that mitigates catastrophic forgetting by penalizing deviations from previously learned model parameters.
    \item \textbf{ER} \cite{chaudhry2019tiny}, an Experience Replay baseline with a limit of 1000 trajectories in the replay buffer; 
    \item \textbf{LOTUS} \cite{wan2024lotus}, a hierarchical imitation learning approach that integrates experience replay with open-vocabulary visual representation models for continuous discovery.
    \item \textbf{M2Distill} \cite{roy2024m2distill}, a multi-modal distillation-based method for lifelong imitation learning that preserves a consistent latent space across vision, language, and action distributions throughout the learning process.
\end{itemize}

\subsection{Quantitative Results}
Table  \ref{experiment:main} provides a comprehensive evaluation of T2S against baseline methods in simulation scenarios. It demonstrates that T2S consistently surpasses most baseline methods across the three evaluation metrics. Specifically, our method exhibits notable effectiveness in mitigating catastrophic forgetting within the LIBERO-GOAL and LIBERO-SPATIAL task suites, achieving reductions in NBT by \(21\%\) and \(11\%\), respectively, effectively resulting in zero forgetting. In the LIBERO-OBJECT suite, increased variability in task layouts introduces additional challenges for visual perception, marginally reducing the AUC. Nevertheless, T2S still consistently outperforms the state-of-the-art baseline, M2Distill, in terms of robustness against forgetting.

\begin{figure}[h]
    \centering
    \includegraphics[width=0.45\textwidth]{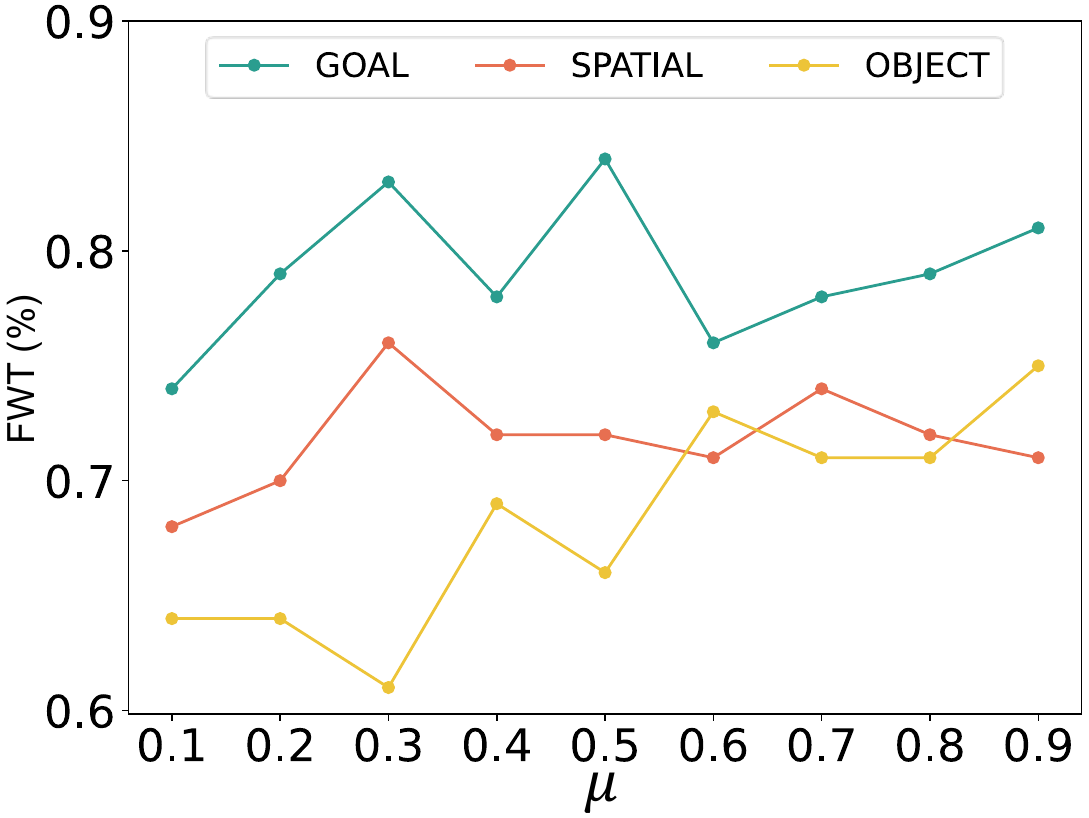}
    \caption{The FWT of T2S for different \(\mu\) values, including LIBERO-GOAL, LIBERO-SPATIAL, and LIBERO-OBJECT task suits, analyzing the impact of \(\mu\) on different types of knowledge transfer.}
    \label{ablation:knowledge}
\end{figure}

\begin{figure}[h]
    \centering
    \includegraphics[width=0.45\textwidth]{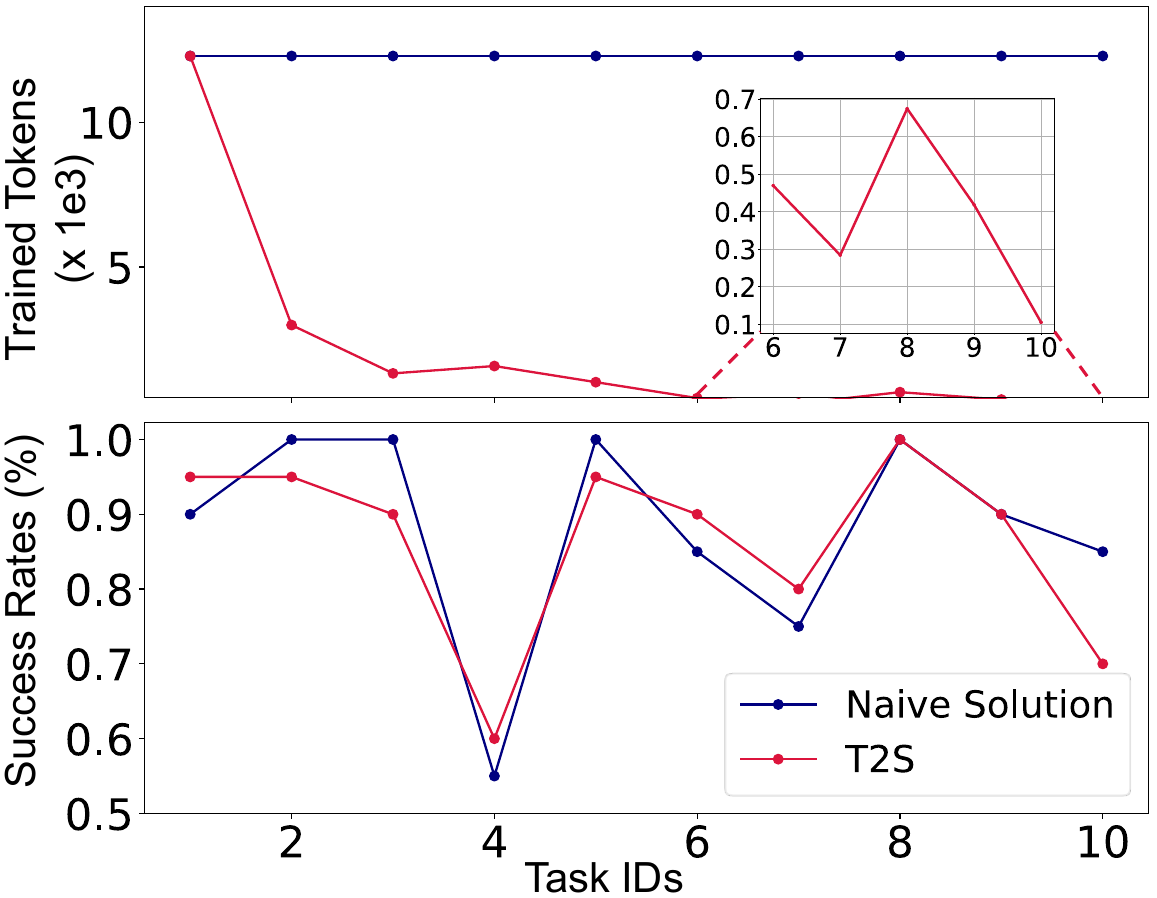}
        \caption{We compare T2S with the naive solution in terms of the number of trainable tokens (for each task) and success rate, showing significantly fewer trainable parameters with little difference in success rates across tasks.}
        \label{ablation:tokens}
\end{figure}

\subsection{Ablation Analysis}
\label{subsec:ablation}
\textbf{Knowledge Transfer.} Fig. \ref{ablation:knowledge} illustrates how the allocation of shared and task-specific tokens influences the transfer of different types of knowledge. On one hand, procedural knowledge transfer generally outperforms declarative knowledge across different \(\mu\) values, suggesting that procedural knowledge is more easily shared between tasks. On the other hand, at \(\mu=0.9\), where only \(10\%\) of task-specific tokens are required for each task, T2S still exhibits effective knowledge transfer, demonstrating that tokenized skills can be efficiently shared across tasks.

\textbf{Token Efficiency.} we compare T2S with the naive approach of learning separate tokens for each task on the LIBERO-GOAL task suit. As shown in Fig. \ref{ablation:tokens}, the top half displays the number of tokens required for training each task in both methods, while the bottom half presents the success rate for each task. T2S can benefit from shared tokens selected from the token pool under the guidance of task instruction. Thus, it can be observed that the number of trainable parameters required for each task is reduced heavily, while achieving a comparable success rate. It indicates that the proposed method can effectively leverage shared knowledge across tasks and minimize the storage burden.

\textbf{Semantic Guidance.} To evaluate the effectiveness of language-guided token selection, we perform an ablation study comparing it with a simpler baseline that activates tokens based solely on task IDs. As shown in Table \ref{taskid}, although task IDs can distinguish different tasks, they lack semantic information regarding object categories and action intents. Consequently, this approach results in significantly lower success rates compared to natural language descriptions. These findings underscore the importance of semantically rich prompts for guiding token selection and enabling effective parameter shearing across tasks.

\begin{table}[t]
\centering
\centering
\caption{Performance of the different language guidance, all metrics are measured based on success rate ($\%$).}
\resizebox{0.45\textwidth}{!}{
\begin{tblr}{
  cells = {c},
  cell{1}{1} = {r=2}{},
  cell{1}{2} = {c=3}{},
  hline{1,5} = {-}{0.08em},
  hline{2} = {2-4}{},
  hline{3} = {-}{},
}
Methods  & LIBERO-GOAL &     &     \\
         & FWT$(\nearrow)$          & NBT$(\searrow)$ & AUC$(\nearrow)$  \\
Language &\textbf{84.0 ($\pm$ 1.0)} &-1.0 ($\pm$ 2.0) &\textbf{70.0 ($\pm$ 2.0)}    \\
TASK IDs &75.7($\pm$ 1.0)            &\textbf{-2.0 ($\pm$ 1.0)}     & 67.2($\pm$ 2.0)

\end{tblr}
}
\label{taskid}
\end{table}

\textbf{Task Order Robustness.} To assess the sensitivity of our method to task ordering, we conduct experiments by randomly shuffling the original task sequence and repeating training five times. As shown in Table~\ref{tab2}, the forward transfer (FWT) varies depending on the informativeness of the initial tasks. However, the forgetting rate (NBT) remains consistently low across different permutations. This robustness indicates that the proposed language-guided token sharing mechanism effectively preserves previously acquired knowledge, allowing the system to maintain stability under varying task orders.

\begin{table}[t]
\centering
\caption{Performance of the different task orders, all metrics are measured based on success rate ($\%$).}
\resizebox{0.45\textwidth}{!}{
\begin{tblr}{
  cells = {c},
  cell{1}{1} = {r=2}{},
  cell{1}{2} = {c=3}{},
  hline{1,5} = {-}{0.08em},
  hline{2} = {2-4}{},
  hline{3} = {-}{},
}
Methods  & LIBERO-GOAL &     &     \\
         & FWT$(\nearrow)$          & NBT$(\nearrow)$  & AUC$(\searrow)$  \\
Default &\textbf{84.0 ($\pm$ 1.0)} &-1.0 ($\pm$ 2.0) &\textbf{70.0 ($\pm$ 2.0)}     \\
Shuffle &80.0 ($\pm$ 1.0)             &-1.0 ($\pm$ 1.0)     &69.6 ($\pm$ 1.0)     
\end{tblr}
}
\label{tab2}
\end{table}


\section{Conclusion}
We propose the Tokenized Skill Scaling (T2S) framework for lifelong robot manipulation tasks, focusing on mitigating catastrophic forgetting and enabling new skill scaling. To achieve this, we transform parameter linear mapping in the traditional transformer to tokenize all parameters to tokens, which allows the model to scale easily through the extension of new tokens. Additionally, we introduce a language-guided skill scaling method to address the issue of linearly growing parameters and to facilitate knowledge sharing across tasks. Through quantitative evaluation of the LIBERO task suites, we demonstrate that our proposed method significantly outperforms baseline approaches in mitigating catastrophic forgetting. 


\bibliographystyle{ieeetr}
\bibliography{references}
\end{document}